\documentclass[10pt,letterpaper]{article}
\usepackage[top=0.85in,left=2.75in,footskip=0.75in,marginparwidth=2in]{geometry}

\usepackage[utf8]{inputenc}

\usepackage{cite}

\usepackage{nameref,hyperref}

\usepackage[right]{lineno}

\usepackage{microtype}
\DisableLigatures[f]{encoding = *, family = * }

\raggedright
\usepackage{changepage}

\usepackage[aboveskip=1pt,labelfont=bf,labelsep=period,singlelinecheck=off]{caption}

\makeatletter
\renewcommand{\@biblabel}[1]{\quad#1.}
\makeatother

\usepackage{lastpage,fancyhdr,graphicx}
\usepackage{epstopdf}
\fancyheadoffset[L]{2.25in}
\fancyfootoffset[L]{2.25in}

\usepackage{color}

\definecolor{gray}{gray}{0.25}

\usepackage{graphicx}

\usepackage{sidecap}

\usepackage{wrapfig}
\usepackage[pscoord]{eso-pic}
\usepackage[fulladjust]{marginnote}
\reversemarginpar

\definecolor{purple}{RGB}{128, 0, 128}
\definecolor{caption_main}{gray}{0.25}
\definecolor{caption_sub}{gray}{0.25}
\usepackage{rotating}
\usepackage{wrapfig}
\newcommand{\etal}{\textit{et al}. }
\newcommand{\ie}{\textit{i}.\textit{e}. }
\newcommand{\eg}{\textit{e}.\textit{g}. }
\newcommand{\citep}{\cite}
\begin{document}
\vspace*{0.35in}
\begin{flushleft}
{\Large
\textbf\newline{On-field player workload exposure and knee injury risk monitoring via deep learning}
}
\newline
\\
William~R.~Johnson\textsuperscript{1,*},
Ajmal~Mian\textsuperscript{2},
David~G.~Lloyd\textsuperscript{3},
Jacqueline~A.~Alderson\textsuperscript{1,4}
\\
\bigskip
\bf{1} {School of Human Sciences (Exercise and Sport Science), The University of Western Australia, Perth, Australia.} 
\\
\bf{2} {Department of Computer Science and Software Engineering, The University of Western Australia, Perth, Australia.} 
\\
\bf{3} {Menzies Health Institute Queensland, and the School of Allied Health Sciences, Griffith University, Gold Coast, Australia.}
\\
\bf{4} {Sports Performance Research Institute New Zealand (SPRINZ), Auckland University of Technology, Auckland, New Zealand.}
\\
\bigskip
* {\href{mailto:bill.johnson@uwa.edu.au}{\color{blue}bill.johnson@uwa.edu.au}\par
\vskip 6pt
\href{https://doi.org/10.1016/j.jbiomech.2019.07.002}{\texttt{\color{blue}https://doi.org/10.1016/j.jbiomech.2019.07.002}}}
\end{flushleft}
\section*{Abstract}
In sports analytics, an understanding of accurate on-field 3D knee joint moments (KJM) could provide an early warning system for athlete workload exposure and knee injury risk. Traditionally, this analysis has relied on captive laboratory force plates and associated downstream biomechanical modeling, and many researchers have approached the problem of portability by extrapolating models built on linear statistics. An alternative approach would be to capitalize on recent advances in deep learning. In this study, using the pre-trained CaffeNet convolutional neural network (CNN) model, multivariate regression of marker-based motion capture to 3D KJM for three sports-related movement types were compared. The strongest overall mean correlation to source modeling of 0.8895 was achieved over the initial 33~\% of stance phase for sidestepping.  The accuracy of these mean predictions of the three critical KJM associated with anterior cruciate ligament (ACL) injury demonstrate the feasibility of on-field knee injury assessment using deep learning in lieu of laboratory embedded force plates.  This multidisciplinary research approach significantly advances machine representation of real-world physical models with practical application for both community and professional level athletes.\\
\bigskip
\noindent\textbf{Keywords} Biomechanics $\cdot$ Wearable sensors $\cdot$ Computer vision $\cdot$ Motion capture $\cdot$ Sports analytics\par
\vskip 24pt
\marginpar{Supplementary material available (\href{http://digitalathlete.org}{\texttt{\color{blue}digitalathlete.org}}).}
\newpage
\section{Introduction}
It is currently not possible to accurately estimate joint loads on the sporting field as the process to estimate these forces and moments generally requires high-fidelity multidimensional force plate inputs and complex biomechanical modeling procedures, traditionally available only in biomechanics laboratories.  The sports biomechanist must instead trade the ecological validity of field-based data capture and manage the limitations of the artificial laboratory environment to accurately model internal and external musculoskeletal loads \citep{4RN387chiari,4RN409elliott}.\par
One of the most devastating sport injuries is the rupture of the anterior cruciate ligament (ACL) which can be a season or career-ending event for the professional athlete \citep{4RN109dallalana}.  Most ACL incidents in team sports such as basketball and hockey are non-contact events (51 to 80~\%), with more than 80~\% reportedly occurring during sidestepping or single-leg landing maneuvers \citep{4RN479donnelly,4RN472shimokochi}. These statistics highlight that the ACL injury mechanism should be generally regarded as an excessive load-related event, resulting specifically from an individual's neuromuscular strategy and associated motion, which by definition is preventable.  Alongside technique factors, studies have identified increased knee joint moments (KJM), specifically high external knee abduction moments during unplanned sidestepping tasks, as a strong indicator of ACL injury risk \citep{4RN703besier,4RN499dempsey}, and as such, real time accurate estimates of ground reaction forces and moments (GRF/M) and knee joint loads could be used as an early warning system to prevent on-field non-contact knee trauma.\par
The orthodox approach to calculating joint loads requires directly recording forces applied to the athlete by either (1) modifying the laboratory environment to better mimic the specific sport requirements, or (2) instrumenting the athlete directly with force transducers or other surrogate wearable sensors to estimate these forces.  In soccer, Jones \etal \citep{4RN469jones} brought the field into the laboratory by mounting turf on the surface of the force plate.  Conversely, Yanai \etal \citep{4RN489yanai} employed the reverse approach by embedding force plates directly into the baseball pitching mound. Alternatively, two main types of wearable sensor technologies have also been used to estimate GRF/M, first, in-shoe pressure sensors \citep{4RN562burns,4RN457liu}, and more recently, body mounted inertial sensors \citep{4RN504karatsidis,4RN592pouliotlaforte,4RN604wundersitz}. Unfortunately, the accuracy of these methods is restricted to simple gait motion (\eg walking), whereby they estimate only a single force component (primarily vertical $F_z$), or the sensor itself (location or added mass) adversely affects performance \citep{4RN562burns,4RN457liu,4RN504karatsidis,4RN592pouliotlaforte,4RN604wundersitz}. 
The current generation of wearable sensors are limited by low-fidelity, low resolution, or uni-dimensional data analysis (\eg velocity) based on gross assumptions of linear regression, which overfit to a simple movement pattern or participant cohort \citep{4RN708camomilla}, however, researchers have reported success deriving kinematics from these devices for movement classification \citep{4RN722pham2018,4RN797watari2018}.
To improve on these methods, a number of research teams have sought to leverage computer vision and data science techniques, and while initial results appear promising, to date they lack validation to ground truth data, or relevance to specific sporting related tasks \citep{4RN493chen,4RN491soopark,4RN492wei}.  For example, Fluit \etal \citep{4RN576fluit} and Yang \etal \citep{4RN607yang} derive GRF/M from motion capture.  However, the former requires the complexity of a full body musculoskeletal model, while the latter again predicts only $F_z$.  These examples of data science and machine learning solutions have relied on basic neural networks and small sample sizes, which means perhaps the biggest untapped opportunity for biomechanics research and practical application is to approach these problems by building on the success of more recent deep learning techniques, which are better suited to exploit large amounts of historical biomechanical data \citep{4RN565choi,4RN468oh,4RN595richter,4RN478sim,4RN755aljaaf}.\par
In the biomechanics laboratory, retro-reflective motion capture is considered the gold standard in marker-based motion analysis, utilizing high-speed video cameras (up to $2,000~Hz$) with built-in strobes of infrared (IR) light to illuminate small spherical retro-reflective passive markers attached to the body \citep{4RN387chiari,4RN399lloyd}. Often captured concurrently with the motion data are analog outputs from force plates providing synchronized GRF/M. The three orthogonal force and moment components recorded are: horizontal (shear) forces $F_x$ and $F_y$, the vertical force $F_z$, and the three rotation moments $M_x$, $M_y$ and $M_z$ about the corresponding force axes.  Force plates can be affected by a variety of systematic errors and installation must be carried out in such a manner as to minimize vibration, and with regard to the frequency and absolute force magnitude of the captured movement.  This means that mounting the plate flush with a concrete floor pad during laboratory construction produces optimal force recordings.  However, this makes the force plate difficult to move or install in outdoor sporting environments, and errors can also be propagated from failures in maintenance, calibration, and operation \citep{4RN593psycharakis,4RN712collins}.\par
Motion and force plate data are conventionally used as inputs to derive the corresponding joint forces and moments via inverse dynamics analysis \citep{4RN400manal,4RN605yamaguchi}.  Over the past twenty years at The University of Western Australia (UWA), upper and lower body biomechanical models have been developed in the scripting language BodyBuilder (Oxford Metrics, Oxford, UK), with the aim of providing repeatable kinematic and kinetic data outputs (\eg KJM) between the three on campus biomechanics laboratories and external research partners \citep{4RN385besier,4RN501chin2010}.  This paper aims to leverage this legacy UWA data collection by using non-linear data science techniques to accurately predict KJM directly from motion capture alone.\par
Deep learning is a branch of machine learning based on the neural network model of the human brain and which uses a number of hidden internal layers \citep{4RN717lecun}.  Enabled by recent increases in computing power, the technique has gained popularity as a powerful new tool in computer vision and natural language processing, and one potentially well-suited to the 3D time-based data structures found in biomechanics \citep{4RN453krizhevsky}.  Caffe (Convolutional Architecture for Fast Feature Embedding), maintained by Berkeley AI Research (BAIR), is one of a growing number of open-source deep learning frameworks \citep{4RN579jia}, alongside others including TensorFlow and Torch \citep{4RN559abadi,4RN567collobert}. Caffe originated from the ImageNet Large Scale Visual Recognition Challenge (2012), is optimized for both CPU and GPU operation, and allows models to be constructed from a library of modules, including convolution (convolutional neural network, CNN) and pooling layers, which facilitates a variety of deep learning approaches \citep{4RN578ichinose,4RN453krizhevsky}. Training a deep learning model from scratch can require a large number of data samples, processing power and time. Fine-tuning (transfer learning) is a technique commonly employed to take an existing related model and only re-train certain higher level components, thus needing relatively less data, time and computational resources. Pre-trained models such as AlexNet, CaffeNet and GoogLeNet are selected according to their relevance to the higher-level data-set.  CaffeNet, for example, was trained on 1.3~million ImageNet images and 1,000 object classes \citep{4RN601szegedy}.\par
Contrary to the traditionally isolated data capture methods in the sport sciences, what made this investigation possible was access to the UWA data archive.
Using this pooled historical data, the aim of the study was to accurately predict extension/flexion, abduction/adduction and internal/external rotation KJM from marker-based motion capture.
Although this would negate the requirement for embedded force plates and the inverse dynamics modeling process, it is still tied to the laboratory.
However, if successful this work would provide the necessary information to facilitate the next phase of the project, which is to drive multivariate regression models (not just classification) from low-fidelity wearable sensor input, trained from high-fidelity laboratory data, for eventual outdoor use.
It was hypothesized that by mimicking the physics behind inverse dynamics the strongest correlations would be achieved via the double-cascade technique from CaffeNet models which had been pre-trained in the relationship between marker-based motion capture and GRF/M.\par
\section{Methods}
\subsection{Design \& setup}
\begin{figure*}
    \begin{center}
    \includegraphics[width=1.0\linewidth]{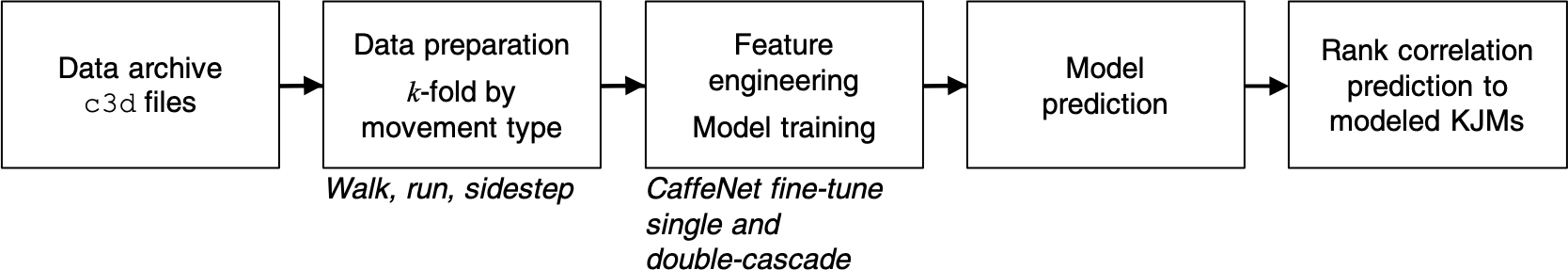}
    \end{center}
    \captionsetup{labelformat=empty}
    \caption{}
    \label{figb}
\end{figure*}

The data used in this new study was captured in multiple biomechanics laboratories over a 17--year period from 2001--2017 (the overall design of the study is shown in Figure~\ref{figb}).  The primary UWA biomechanics laboratory was a controlled space which utilized lights and wall paint with reduced IR properties.
Over this period, the floor surface coverings have varied from short-pile wool carpet squares to artificial turf laid on, and around, the force plate surface.
While not directly tested, all selected surface coverings during this time had negligible underlay or cushioning component in an effort to minimize any dampening characteristics.
It is important to note that the variety of surfaces were spread amongst both training and test sets, enabling the prototype to proceed without surface covering calibration.
However, further calibration of the model would be required for future outdoor use with variant surfaces (\eg grass).
Trials were collected from a young healthy athletic population (male and female, amateur to professional), and with pathological or clinical cohort samples excluded.\par
\begin{wrapfigure}{l}{0.42\textwidth}
    \includegraphics[width=1.0\linewidth]{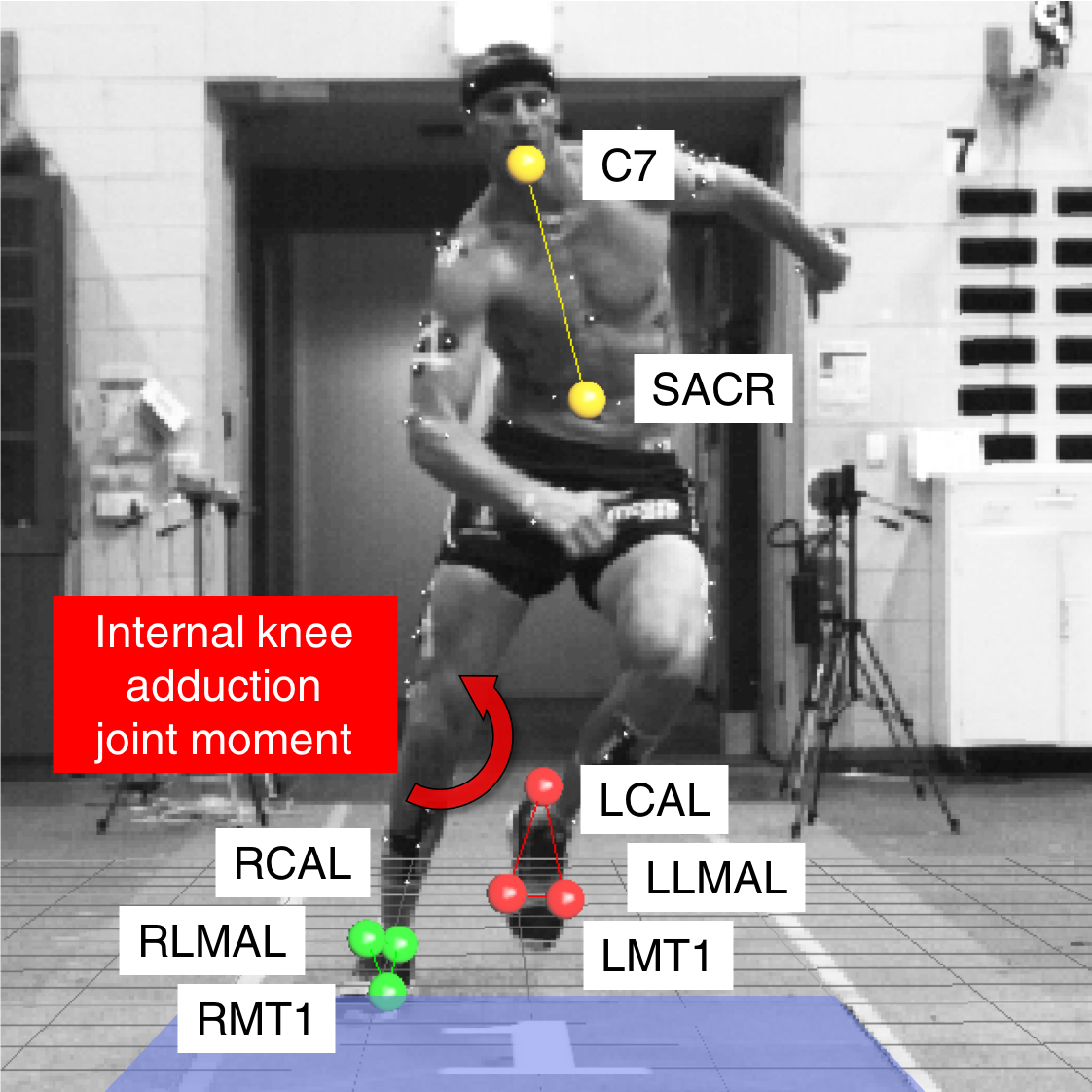}
    \captionsetup{labelformat=empty}
    \caption{}
    \label{figa}
\end{wrapfigure}

\marginpar{
    \begin{flushleft}
    \vspace{-9cm}
    {\color{caption_main}\textbf{Figure \ref{figb}. Study overall \mbox{design}.}}
    \end{flushleft}
}
\marginpar{
    \vspace{7mm} 
    {\color{gray}\textbf{Figure \ref{figa}. Laboratory motion and force plate data capture overlay.}~{\color{caption_sub}The eight labeled markers used are shown artificially colored and enlarged, and the force plate highlighted blue. An internal knee adduction joint moment is depicted.}}
}
Sample trials were collected using 12--20 cameras (Vicon model types MCam2, MX13 and T40S; Oxford Metrics, Oxford, UK) mounted on tripods and wall-brackets that were aimed at the desired 3D reconstruction volume.
AMTI force plates were used to record the six vector GRF/M (Advanced Mechanical Technology Inc., Watertown, MA).
Equipment setup and calibration was conducted to manufacturer specifications using the proprietary software at the time of collection (Workstation v4.6 to Nexus v2.5), with motion and analog data output in the public-domain \lq{coordinate~3D}\rq~\texttt{c3d} binary file format (maintained by Motion Lab Systems, Baton Rouge, LA).
Since the full UWA marker set has evolved to comprise between 24--67 markers, a subset of eight passive retro-reflective markers (cervical vertebra C7; sacrum SACR; plus bilateral hallux MT1, calcaneus CAL, and lateral ankle malleolus LMAL) were selected for the present investigation to maximize trial inclusion, relevance to the movement type, and downstream KJM output (Figure~\ref{figa}) \citep{4RN385besier,4RN499dempsey,4RN490johnson}.\par
\subsection{Data preparation}
Data mining the archive of 458,372 motion capture files was approved by the UWA human research ethics committee (RA/4/1/8415) and no new data capture was undertaken. The data-set contributions by sex, height, and mass were male 62.8~$\%$, female 37.2~$\%$, height $1.766~\pm~0.097~m$, and mass $74.5~\pm~12.2~kg$ respectively; and the depersonalized nature of the samples meant participant-trial membership (the number of participants in relation to the number of trials) was not considered. Data processing was conducted using MATLAB R2017b (MathWorks, Natick, MA) in conjunction with the Biomechanical ToolKit 0.3 \citep{4RN463barre}, Python 2.7 (Python Software Foundation, Beaverton, OR) and R 3.4.3 \citep{4RN494rcore}, running on Ubuntu v16.04 (Canonical, London, UK). Hardware used was a desktop PC, Core i7 4GHz CPU, with 32GB RAM and NVIDIA multi-GPU configuration (TITAN~X \& TITAN~Xp; NVIDIA Corporation, Santa Clara, CA).\par
The preparation phase was necessary to ensure the integrity of the marker trajectories and force plate analog channels, and to limit the impact of manual errors that may have propagated through the original data capture pipeline \citep{4RN587merriaux,4RN593psycharakis}. First, the data mining relied on trials with the eight required marker trajectories being contiguous and labeled, together with associated KJM, and force plate channel data present throughout the entire stance phase.
To validate the model by comparing calculated and predicted KJM over a standardized stance phase, it was necessary for both the training and test sets to include gait event information defined from a common source, which in this instance was the force plate.
This requirement would not extend to real-world use, where it is envisaged that event data would be predicted from a continuous stream of input kinematics from wearable sensors located on the feet \citep{4RN731falbriard2018}.
For this study, foot-strike (FS) was automatically detected by foot position being within the force plate corners, and if calibrated $F_z$ was continuously above the threshold (20~$N$) for a defined period (0.025~$s$); toe-off (TO) by $F_z$ falling below a second threshold (10~$N$) \citep{4RN487milner,4RN589oconnor,4RN602tirosh}.
As the eight chosen markers did not include any on the shank, the traditional kinematic definitions of heel strike and forefoot strike were unavailable.
A custom approach was adopted to demonstrate the spread and variety of different foot segment orientations at foot-strike.
This {\lq{sagittal foot orientation}\rq} was reported by comparing the vertical height of forefoot and primary rearfoot markers in the phase defined by FS and the mean 25--30~\% of stance (within $\pm$~1~\% tolerance), to categorize general foot orientation at contact as heel down (HD), flat (FL), or toe down (TD) (Table~\ref{tab1v2}).
The determination of sagittal foot orientation was used to illustrate the variety of running patterns present in the data, and no judgment of model performance according to foot orientation was made.\par
Cubic spline interpolation was used to time-normalize the marker trajectories from FS minus 66~\% of stance (total 125 samples, typically 0.5~sec at 250~Hz) and force plate channels from FS minus 16~\% of stance (700 samples, correspondingly 0.35~sec at 2,000~Hz), both until TO.
Normalization allowed the model to be agnostic to the original motion capture and force plate data capture frequencies.
Furthermore, our earlier study \citep{4RN490johnson} demonstrated the importance of an additional lead-in period for marker trajectories, and the use of different sample sizes to reflect the relative ratio of the original capture frequencies.
Duplicate and invalid capture samples were automatically removed, with no regard for earlier filtering, and in the case of sidestepping, whether it was performed in a planned or unplanned manner.
If the motion capture and force plate data survived these hygiene tests it was reassembled into the data-set arrays X (predictor samples $\times$ input features) and y (response samples $\times$ output features) typical of the format used by multivariate regression \citep{4RN464chun}.\par
Kinematic templates were used to categorize the three movement types, selected for their progressive complexity and relevance in sports: walking, running, and sidestepping (opposite foot to direction of travel); the threshold for walking to running used was $2.16~m/s$ \citep{4RN707segers}.
A small proportion of sidestepping trials used crossover technique and these were removed to avoid contaminating the model.
KJM were considered only for the stance limb, and the majority of participants reported being right leg dominant (Table~\ref{tab1v2}).\par
\begin{table*}
\begin{adjustwidth}{-1.5in}{0in} 
    \begin{center}
    \fontsize{8.0}{8.0}\selectfont
    \renewcommand{\arraystretch}{1.4}
    \caption[Data-set characteristics, sagittal foot orientation, by movement type and stance limb]{\color{caption_main}\textbf{Data-set characteristics, sagittal foot orientation, by movement type and stance limb.}}
    \vskip 12pt
    \label{tab1v2}
    \begin{tabular}{|c|c|c|c|c|c|}																									
    \hline																									
    Movement	&	Stance	&	Data-set		&	\multicolumn{3}{|c|}{Sagittal foot orientation}																	\\ %
    type	&	limb	&	samples		&	Heel down (HD)					&	Flat (FL)					&	Toe down (TD)					\\ %
    \hline\hline																									
    Walk	&	L	&	570		&	570			(100.0~\%)		&	0			(0.0~\%)		&	0			(0.0~\%)		\\ %
    Walk	&	R	&	646		&	646			(100.0~\%)		&	0			(0.0~\%)		&	0			(0.0~\%)		\\ %
    \hline																									
    Run	&	L	&	233		&	209			(89.7~\%)		&	23			(9.9~\%)		&	1			(0.4~\%)		\\ %
    Run	&	R	&	884		&	811			(91.7~\%)		&	71			(8.0~\%)		&	2			(0.2~\%)		\\ %
    \hline																									
    Sidestep	&	L	&	566		&	457			(80.7~\%)		&	88			(15.5~\%)		&	21			(3.7~\%)		\\ %
    Sidestep	&	R	&	1527		&	1162			(76.1~\%)		&	325			(21.3~\%)		&	40			(2.6~\%)		\\ %
    \hline
    \end{tabular}
    \end{center}
\end{adjustwidth}
\end{table*}
\begin{figure*}
\begin{adjustwidth}{-2in}{0in}
    \begin{center}
    \includegraphics[width=0.9\linewidth]{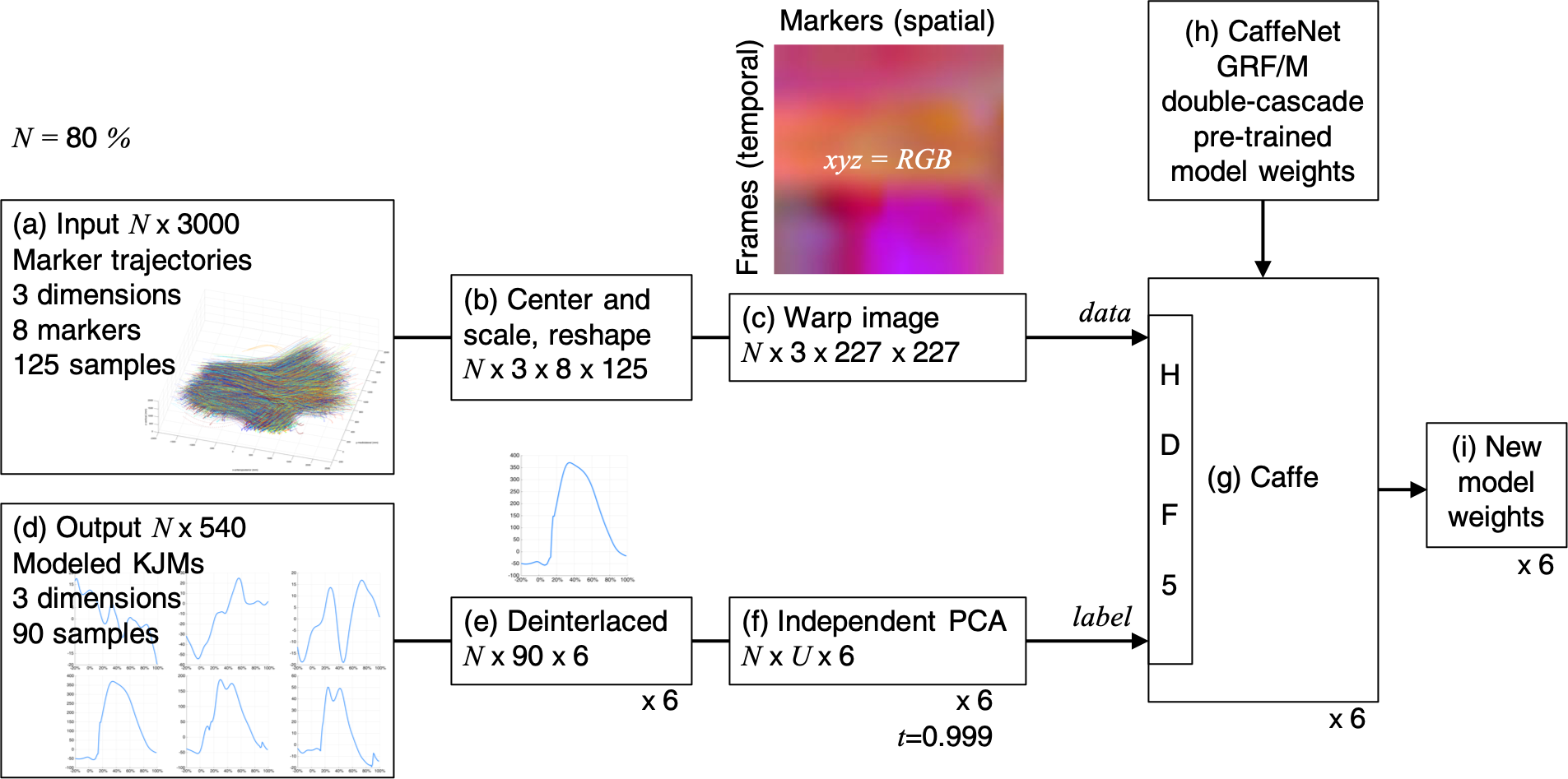}
    \end{center}
    \vspace*{6mm}   
    \caption[CaffeNet CNN model training (double-cascade)]{\color{caption_main}\textbf{CaffeNet CNN model training (double-cascade).}~{\color{caption_sub}Input marker trajectories flattened to color images, output KJM 6-way deinterlaced PCA reduced.}}
    \label{figc}
    \footnotesize{(a) The input marker trajectories for each trial total 3,000 features (columns of data).\\}
    \footnotesize{(b) Input data is centered, scaled, and reshaped, then (c) flattened (warped) into 227~$\times$~227 pixel images.\\}
    \footnotesize{(d) The corresponding modeled KJM output has 540 features.\\}
    \footnotesize{(e) The number of output features is reduced first by deinterlacing (splitting) the KJM into their six component waveforms, each 90 features, then (f) reduced again using PCA ($t$=0.999).  Subsequent processes are executed for each of the six waveforms (x~6).\\}
    \footnotesize{(g) The HDF5 file format (\href{https://www.hdfgroup.org/HDF5/}{\texttt{\color{blue}{hdfgroup.org}}}) is used to pass the large data structures (input `data', output `label') into the CNN.\\}
    \footnotesize{(h) In this example, weights from an earlier CaffeNet GRF/M model are used to help improve the accuracy of the new model (double-cascade).\\}
    \footnotesize{(i) The new model weights are used for KJM prediction from the test-set.}
\end{adjustwidth}
\end{figure*}
\begin{figure*}
\begin{adjustwidth}{-2in}{0in}
    \begin{center}
    \includegraphics[width=0.9\linewidth]{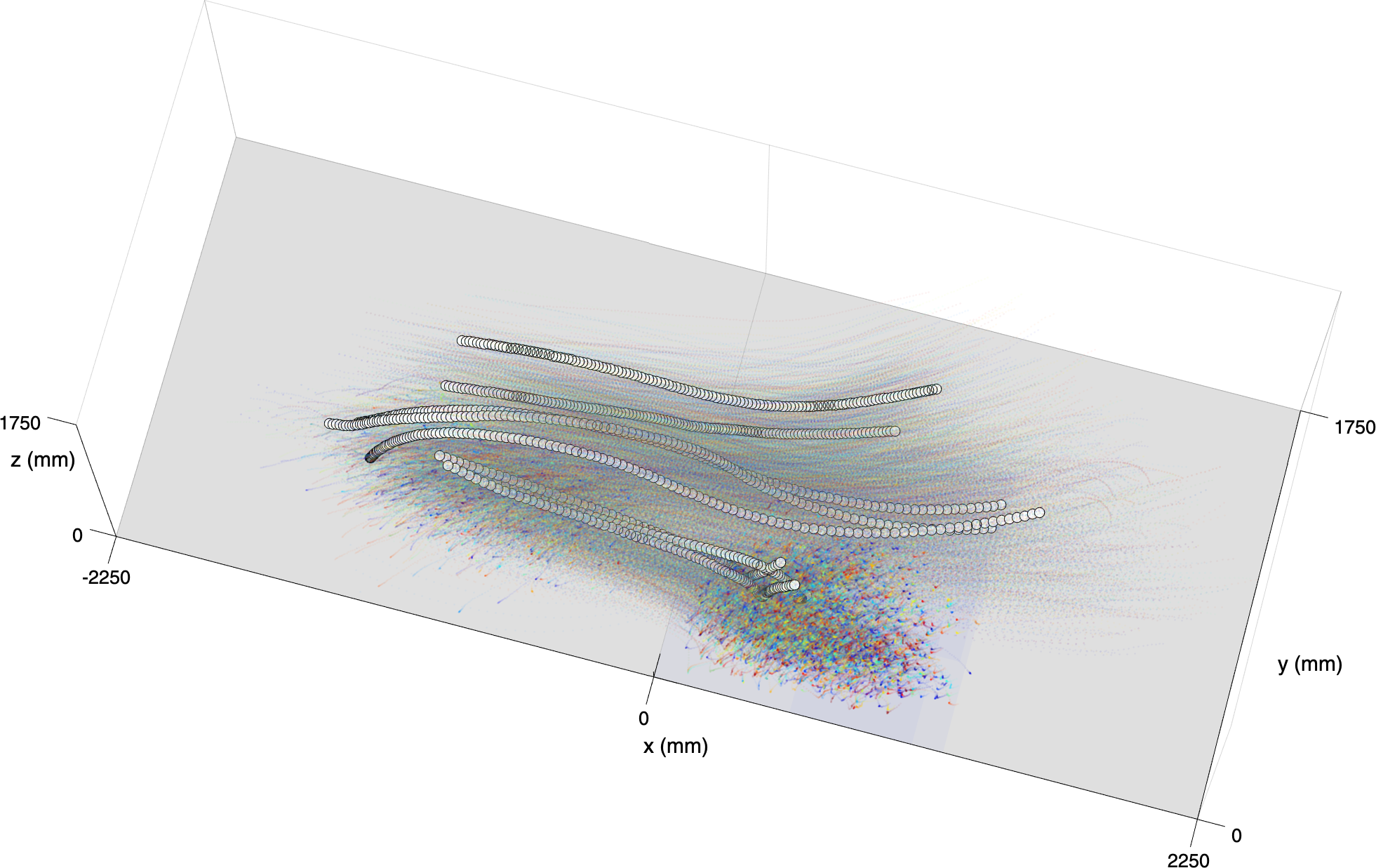}
    \end{center}
    \vspace*{6mm}   
    \caption[Training-set eight marker trajectories, sidestep~left movement type (off right stance limb)]{\color{caption_main}\textbf{Training-set eight marker trajectories, sidestep~left movement type (off right stance limb).}~{\color{caption_sub}Combined 1,222 predictor samples (80~\% of 1,527), viewed as a conventional 3D volume space and with one trial highlighted.}}
    \label{figfa}
\end{adjustwidth}
\end{figure*}
\subsection{Feature engineering \& model training}
\begin{wrapfigure}{l}{0.46\textwidth}
    \includegraphics[width=1.0\linewidth]{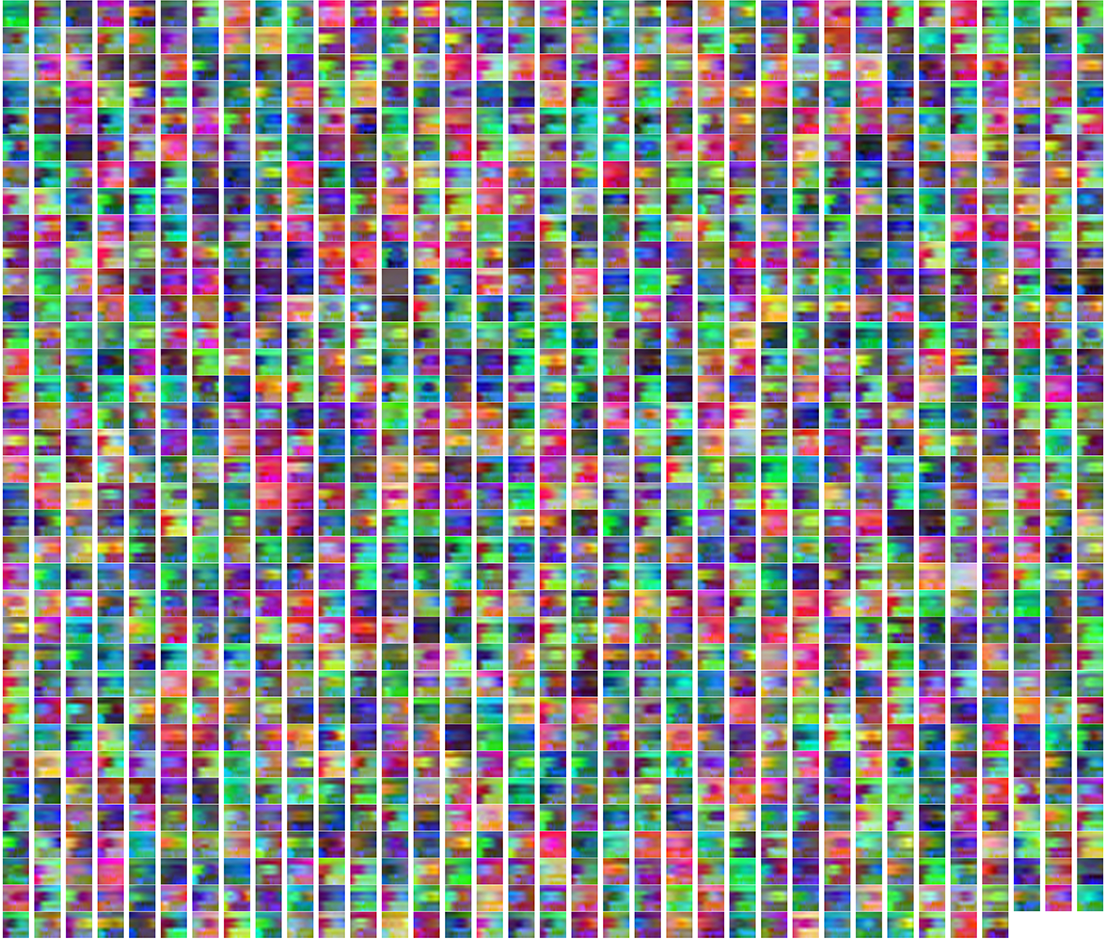}
    \captionsetup{labelformat=empty}
    \caption{}
    \label{figfb}
\end{wrapfigure}

The Caffe deep learning framework was used to fine-tune the native CaffeNet model (Figure~\ref{figc}).
However, to be presented to CaffeNet and benefit from the model's pre-training on the large-scale ImageNet database, the study training-set input marker trajectories needed to be converted from their native spatio-temporal state to a set of static color images \citep{4RN574du,4RN582ke} (Figures~\ref{figfa}~\&~\ref{figfb}).
This was achieved by mapping the coordinates of each marker ($x$, $y$, $z$) to the image additive color model ($R$, $G$, $B$), the eight markers to the image width, and the 125 samples to the image height.
The resultant 8~$\times$~125 pixel image was warped to 227~$\times$~227 pixels to suit CaffeNet using cubic spline interpolation (input: predictor samples~$\times$~227~$\times$~227).\par
\marginpar{
    \begin{flushleft}
    \vspace{.7cm}
    {\color{caption_main}\textbf{Figure \ref{figfb}. The spatio-temporal marker trajectories in Figure~\ref{figfa} were presented to the CNN as 1,222 individual color images.}}
    \end{flushleft}
}
A common technique in CNN architecture to improve performance is to minimize the number of output features. The training-set output KJM data was deinterlaced into its six component waveforms (internal moments $LKJM_x$, $LKJM_y$, $LKJM_z$, $RKJM_x$, $RKJM_y$, and $RKJM_z$), each of which was then further reduced using Principal Component Analysis (PCA). PCA with a tuning threshold $t$=0.999 was selected for its compression accuracy with this data type \citep{4RN591pearson}, 
which for example for the sidestep (right stance limb) resulted in $RKJM_z$ (internal/external rotation moment) reducing from 90 to 59 features (output: response samples~$\times$~59).
The PCA compression was repeated for each of the six deinterlaced waveforms.
Rather than the native CaffeNet object classification, the fine-tuned model was required to perform multivariate regression, \ie produce waveform outputs. The fine-tuning process by definition requires the CNN architecture to exactly match that used during original training, with the exception of allowing the final loss branches to be sliced off and replaced with another, exactly for such purposes as required by this study. The CaffeNet architecture was modified by replacing the final 1,000-dimensional SoftMax layer with a Euclidean loss layer (with dimensions as defined dynamically by the output downsampling) thereby converting it from a classification network into a multivariate regression network.\par
Fine-tuning deep learning networks allows smaller downstream sample sizes to leverage weighting relationships built on earlier training at scale \citep{4RN758zamir}. In the current study, new model training was carried out either via a single fine-tune or a double-cascade from pre-trained CNN models. The single fine-tune models were created from the CaffeNet source, itself pre-trained on the ImageNet database, and then trained on the KJM training-set. The double-cascade models were created from variants of the CaffeNet model, pre-trained using an earlier GRF/M training-set, and subsequently trained on the KJM training-set (\ie fine-tuned twice). The donor weights for the double-cascade were provided by the strongest GRF/M model (single fold, 33~\% stance, sidestepping, right stance limb) selected from earlier prototypes.\par
The study training and test-sets were formed from a random shuffle and split of the complete data-set of marker-based motion capture input (predictor) to KJM output (response). As per data science convention, the majority of results were derived from a single 80:20 fold, however early criticism suggested that the model was overfitting to this one fold \citep{4RN467domingos}. Therefore, within time and resource constraints, regression for one movement type was tested over 5-folds, and because of its relevance to knee injury risk but also being the movement type with the largest number of samples, the sidestep movement (right stance limb) was selected for this additional investigation. Testing over 5-folds was achieved with 80:20 splits whereby each member of the data-set was guaranteed to be a participant of each of the five test-sets only once.  The mean of the 5-folds experiments was then compared with the single fold results. The prediction occurred over the initial 33~\% of a time normalized stance phase, the period selected for its relevance to injury risk, and was reported for the three categorized sports-related movement types: walking, running, and sidestepping.
The precision of the outcome of all the experiments was assured by comparing the six vector KJM predicted by the CaffeNet regression model (fine-tune or double-cascade) with those calculated by inverse dynamics using both correlation coefficient $r$ and $rRMSE$ \citep{4RN713ren}.\par
\section{Results}
Of the single fine-tune investigations, the strongest mean KJM correlation was found for the left stance limb during sidestepping $r(LKJM_{mean})$ 0.9179 (shown bolded, Table~\ref{tab2v2}), and the weakest for the right stance limb also during sidestepping $r(RKJM_{mean})$ 0.8168.\par
Applying the double-cascade technique caused all but one correlation to improve compared with the single fine-tune, and also the league table of results to reorder. The strongest mean correlation remained the left stance limb in sidestepping $r(LKJM_{mean})$ 0.9277 (bolded, Table~\ref{tab3v2}), a rise of $+1.1~\%$, and with individual contributing components $r(LKJM_x)$ extension/flexion 0.9829, $r(LKJM_y)$ abduction/adduction 0.9050, $r(LKJM_z$) internal/external rotation 0.8953. However, the greatest improvement from the double-cascade was observed for the right stance limb in sidestepping, for which the earlier $r(RKJM_{mean})$ 0.8168 was improved to $r(RKJM_{mean})$ 0.8512, a significant increase of $+4.2~\%$ ($p~<~0.01$ \citep{4RN756mann}) over the single fine-tune, from components $r(RKJM_x)$ extension/flexion 0.9865, $r(RKJM_y)$ abduction/adduction 0.8368, $r(RKJM_z$) internal/external rotation 0.7304.
The double-cascade technique resulted in a mean increase of $+1.8~\%$, and the sidestepping pair (average of both stance limbs) provided the strongest overall mean correlation $r(KJM_{mean})$ 0.8895.\par
\begin{figure*}
\begin{adjustwidth}{-2in}{0in}
    \begin{center}
    \includegraphics[width=1.0\linewidth]{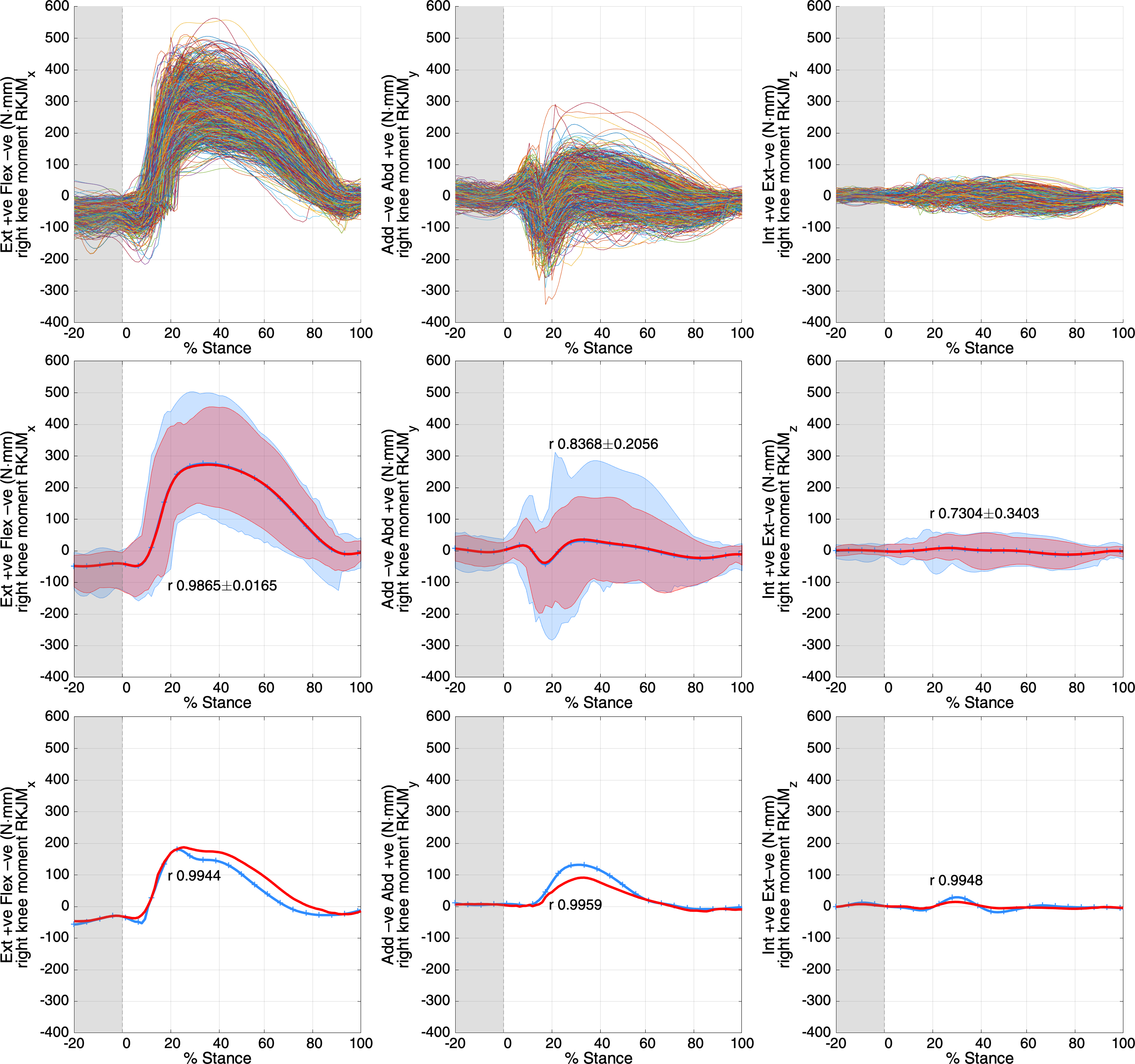}
    \end{center}
    \vspace*{6mm}
    \caption[Ground truth versus predicted response]{\color{caption_main}\textbf{Ground truth versus predicted response.}~{\color{caption_sub}Top, training-set modeled output sidestep~left (right stance foot) internal moments $RKJM_x$ extension/flexion, $RKJM_y$ abduction/adduction, and $RKJM_z$ internal/external rotation, 1,222 samples versus stance phase.  Middle and lower, test-set modeled internal KJM (blue, ticks), and predicted response (red), using CaffeNet double-cascade, 6-way deinterlaced, correlations over initial 33~\% stance phase.  Middle, min/max range and mean predicted response; lower, individual sample with the strongest $r(RKJM_{mean})$.}}
    \label{figd}
\end{adjustwidth}
\end{figure*}

For having the largest number of samples, sidestepping (right stance limb) was investigated further. First, by cross-validation over five k-folds, for which the similarity of the average correlation $r(RKJM_{mean})$ 0.8472 compared with the single fold analysis 0.8512 indicated overfitting had been avoided. Second, by illustration, the output KJM training-set, the test-set predicted response min/max range and mean, and the comparison for the corresponding test sample with the strongest $r(RKJM_{mean})$ (Figure~\ref{figd}).\par
\begin{table*}
\begin{adjustwidth}{-1.5in}{0in} 
    \begin{center}
    \fontsize{8.0}{8.0}\selectfont
    \renewcommand{\arraystretch}{1.4}
    \caption[KJM component vector and mean correspondence, CNN single fine-tune]{\color{caption_main}\textbf{KJM component vector and mean correspondence $r(rRMSE~\%)$, CNN single fine-tune.}~{\color{caption_sub}33~\%~stance, by movement type and stance limb. CaffeNet CNN, output 6-way deinterlaced PCA reduced, single 80:20 fold.}}
    \vskip 12pt
    \label{tab2v2}
    \begin{tabular}{|c|c|c|c|c|c|}																									
    \hline																									
    Movement	&	Stance	&	$KJM_x$~Ext/Flex &	$KJM_y$~Add/Abd	&	$KJM_z$~Int/Ext	& $KJM_{mean}$		\\ %
    type    	&	limb	&	$r(rRMSE~\%)$	 &	$r(rRMSE~\%)$	&	$r(rRMSE~\%)$	& $r(rRMSE~\%)$		\\ %
    \hline\hline																									
    Walk	&	L	&	0.9843	(16.0)	&	0.9672	(16.2)	&	0.6462	(31.8)	&	0.8659	(21.3)      	\\ 
    Walk	&	R	&	0.9860	(13.8)	&	0.9389	(16.4)	&	0.6762	(30.3)	&	0.8670	(20.2)      	\\ 
    \hline																									
    Run	&	L	&	0.9807	(13.5)	&	0.8602	(25.3)	&	0.7733	(27.2)	&	0.8714	(22.0)          	\\ 
    Run	&	R	&	0.9905	(7.8)	&	0.7210	(31.7)	&	0.7430	(24.3)	&	0.8182	(21.3)	            \\ 
    \hline																									
    Sidestep	&	L	&	0.9815	(8.3)	&	0.8854	(16.6)	&	0.8869	(16.2)	&	\textbf{0.9179}	(13.7)	\\ 
    Sidestep	&	R	&	0.9838	(9.7)	&	0.7930	(24.1)	&	0.6736	(28.2)	&	0.8168	(20.7)      	\\ 
    \hline
    \end{tabular}
    \end{center}
\end{adjustwidth}
\end{table*}
\begin{table*}
\begin{adjustwidth}{-1.5in}{0in} 
    \begin{center}
    \fontsize{8.0}{8.0}\selectfont
    \renewcommand{\arraystretch}{1.4}
    \caption[KJM component vector and mean correspondence, CNN double-cascade]{\color{caption_main}\textbf{KJM component vector and mean correspondence $r(rRMSE~\%)$, CNN double-cascade.}~{\color{caption_sub}33~\%~stance, by movement type and stance limb. CaffeNet CNN, output 6-way deinterlaced PCA reduced, single 80:20 fold.}}
    \vskip 12pt
    \label{tab3v2}
    \begin{tabular}{|c|c|c|c|c|c|c|}																									
    \hline																									
    Movement	&	Stance	&	$KJM_x$~Ext/Flex &	$KJM_y$~Add/Abd	&	$KJM_z$~Int/Ext	& $KJM_{mean}$	&	Improvement~(r~\%)\\ %
    type    	&	limb	&	$r(rRMSE~\%)$	 &	$r(rRMSE~\%)$	&	$r(rRMSE~\%)$	& $r(rRMSE~\%)$	&	over single fine-tune	  \\ %
    \hline\hline																									
    Walk	&	L	&	0.9848	(13.7)	&	0.9645	(16.3)	&	0.7008	(30.0)	&	0.8834	(20.0)	&	+2.0                  \\ 
    Walk	&	R	&	0.9900	(12.1)	&	0.9350	(16.6)	&	0.6604	(31.7)	&	0.8618	(20.1)	&	-0.6	              \\ 
    \hline																									
    Run	&	L	&	0.9768	(13.8)	&	0.8657	(23.2)	&	0.7841	(24.7)	&	0.8756	(20.6)	&	+0.5	                  \\ 
    Run	&	R	&	0.9915	(7.3)	&	0.7635	(29.5)	&	0.7934	(22.5)	&	0.8495	(19.8)	&	+3.8	                  \\ 
    \hline																									
    Sidestep	&	L	&	0.9829	(8.0)	&	0.9050	(15.4)	&	0.8953	(16.0)	&	\textbf{0.9277}	(13.2)	&	+1.1	\\ 
    Sidestep	&	R	&	0.9865	(8.7)	&	0.8368	(21.7)	&	0.7304	(26.2)	&	0.8512	(18.9)	&	+4.2	        \\ 
    \hline
    \end{tabular}
    \end{center}
\end{adjustwidth}
\end{table*}
\section{Discussion}
Although the uptake of deep learning continues to increase across all disciplines, examples of modeling biomechanics with CNN fine-tuning are rare with researchers preferring to explore linear models, or attempting to train deep learning models from scratch. This study provides an end-to-end example which illustrates the process to repackage sports biomechanics data (flattening spatio-temporal input, dimensionality reducing output), and modify CNN architecture (replacing SoftMax with Euclidean final loss layer), to leverage existing big data for new biomechanics applications.\par
Using the pre-trained CaffeNet CNN model, multivariate regression of marker-based motion capture to 3D KJM for three sports-related movement tasks were compared. Sidestepping recorded the strongest correlations to source modeling, individually for the left stance limb $r(LKJM_{mean})$ 0.9277, and overall with both stance limbs combined $r(KJM_{mean})$ 0.8895.  The minimum relationship was found with right stance limb running $r(RKJM_{mean})$ 0.8495 demonstrating the general high performance of the model. These results were achieved by using a double-cascade approach to fine-tune KJM CNN models from earlier GRF/M model weights. This purely data science technique was able to improve correlations compared with the single fine-tune approach by at most $+4.2~\%$ ($p~<~0.01$) and on average $+1.8~\%$. The double-cascade technique improved results all but one movement case, and thus the hypothesis was supported.\par
The limitations of the study relate to the exposure to systematic and manual errors in the original preparation of: (a) the marker-based motion capture; (b) the recording of analog force plate GRF/M; and (c) traditional KJM modeling. The number of samples available for KJM analysis is lower, ranging from 1,527 for sidestepping (right stance limb) to 233 for running (left stance limb), compared with 2,196 for GRF/M modeling. This difference is reflected in the lower correlation values reported for the KJM models compared with the earlier GRF/M analysis which indicates the marker trajectory to KJM relationship is more challenging for the CNN, particularly for the knee internal/external rotation moment $r(KJM_z)$. One reason the GRF/M models demonstrate excellent accuracy from only 2,196 data-set samples (not millions) is because of the big data inherent to the ImageNet pre-training of the CNN model, but this also provides an indication of the minimum number of samples required for this method to improve. Increasing the number of KJM samples available with each movement type to match or exceed the GRF/M model is expected to result in a concomitant improvement in KJM model accuracy and consistency. Regardless, this is currently the largest biomechanical study in terms of the number of samples, and the time period over which the data was collected.\par
\section{Conclusions}
The accurate estimate of KJM directly from motion capture, without the use of embedded force plates and inverse dynamics modeling procedures, is a novel approach for the biomechanics community. Through a unique combination of deep learning data engineering techniques, this study was able to extract value from legacy biomechanics data otherwise trapped on legacy external hard-drives.
Refining the movement type classification, and using a larger number of trial samples, will improve the relevance and robustness of the model, and serve as a precursor to real-time on-field use.
Building on earlier GRF/M modeling using deep learning driven purely by motion data, the current study predicts KJM, and with significantly improved correlation performance using the double-cascade technique.
These are important and necessary incremental steps for the goal of in-game biomechanical analysis, and which lay the foundation for future work to drive the multivariate regression not from markers, but from a small number of similarly-located accelerometers.
When this is accomplished, relevant, real-time, and on-field loading information will be available.
For the community player, this approach has the potential to disrupt the wearable sensor market by enabling next-generation classification of movement types and workload exposure.  For professional sport, an understanding of on-field KJM could be used to alert coaches, players and medical staff about the real-time effectiveness of training interventions and changes to injury risk.\par
\section{Acknowledgements}
This project was partially supported by the ARC Discovery Grant DP160101458 and an Australian Government Research Training Program Scholarship. NVIDIA Corporation is gratefully acknowledged for the GPU provision through its Hardware Grant Program. Portions of data included in this study have been funded by NHMRC grant 400937.\par
\nolinenumbers

\end{document}